\documentclass[10pt,twocolumn,letterpaper]{article}

 \usepackage[pagenumbers]{cvpr} %

\usepackage[dvipsnames]{xcolor}

\usepackage{multirow}

\definecolor{cvprblue}{rgb}{0.21,0.49,0.74}
\usepackage[pagebackref,breaklinks,colorlinks,citecolor=cvprblue]{hyperref}

\def\paperID{****} %
\def\confName{CVPR}
\def\confYear{2024}

\title{Hardness-Aware Scene Synthesis for Semi-Supervised 3D Object Detection}

\author{Shuai Zeng$^{1}$ \quad Wenzhao Zheng$^{2,3,}$\footnotemark[1] \quad Jiwen Lu$^{2}$\quad Haibin Yan$^{1,}$\footnotemark[2] \\
$^{1}$Beijing University of Posts and Telecommunications, \\
$^{2}$Tsinghua University, $^{3}$UC Berkeley \\
\texttt{\{zengshuai,eyanhaibin\}@bupt.edu.cn;} \\
\texttt{wenzhao.zheng@outlook.com; lujiwen@tsinghua.edu.cn}
}

\begin{document}
 \maketitle
 
 \renewcommand{\thefootnote}{\fnsymbol{footnote}}
\footnotetext[1]{Project leader. $\dagger$Corresponding author.}
\renewcommand{\thefootnote}{\arabic{footnote}}

\begin{abstract}
3D object detection aims to recover the 3D information of concerning objects and serves as the fundamental task of autonomous driving perception.
Its performance greatly depends on the scale of labeled training data, yet it is costly to obtain high-quality annotations for point cloud data.
While conventional methods focus on generating pseudo-labels for unlabeled samples as supplements for training, the structural nature of 3D point cloud data facilitates the composition of objects and backgrounds to synthesize realistic scenes.
Motivated by this, we propose a hardness-aware scene synthesis (HASS) method to generate adaptive synthetic scenes to improve the generalization of the detection models.
We obtain pseudo-labels for unlabeled objects and generate diverse scenes with different compositions of objects and backgrounds.
As the scene synthesis is sensitive to the quality of pseudo-labels, we further propose a hardness-aware strategy to reduce the effect of low-quality pseudo-labels and maintain a dynamic pseudo-database to ensure the diversity and quality of synthetic scenes.
Extensive experimental results on the widely used KITTI and Waymo datasets demonstrate the superiority of the proposed HASS method, which outperforms existing semi-supervised learning methods on 3D object detection.
Code: \url{https://github.com/wzzheng/HASS}.
\end{abstract}

\newcommand{\tablevspace}{\vspace{-5mm}} %
\newcommand{\figvspace}{\vspace{-5mm}} %

\section{Introduction}

3D object detection~\cite{wu2021multiple,zhou2020salient} is an important task for autonomous driving~\cite{tpvformer,genad,surroundocc,pointocc}, yet the high cost of manually labeling data impedes its applications \cite{mao2021one,fang2020augmented,meng2021towards,meng2020weakly,selfocc}. 
To improve the performance when training samples are insufficient, semi-supervised learning methods~\cite{sajjadi2016regularization,bachman2014learning,tarvainen2017mean,rasmus2015semi,berthelot2019mixmatch,sohn2020fixmatch,lee2013pseudo} aim to utilize additional unlabeled samples to improve the generalization performance of the model.
\begin{figure}[tb]
\centering
\includegraphics[width=0.475\textwidth]{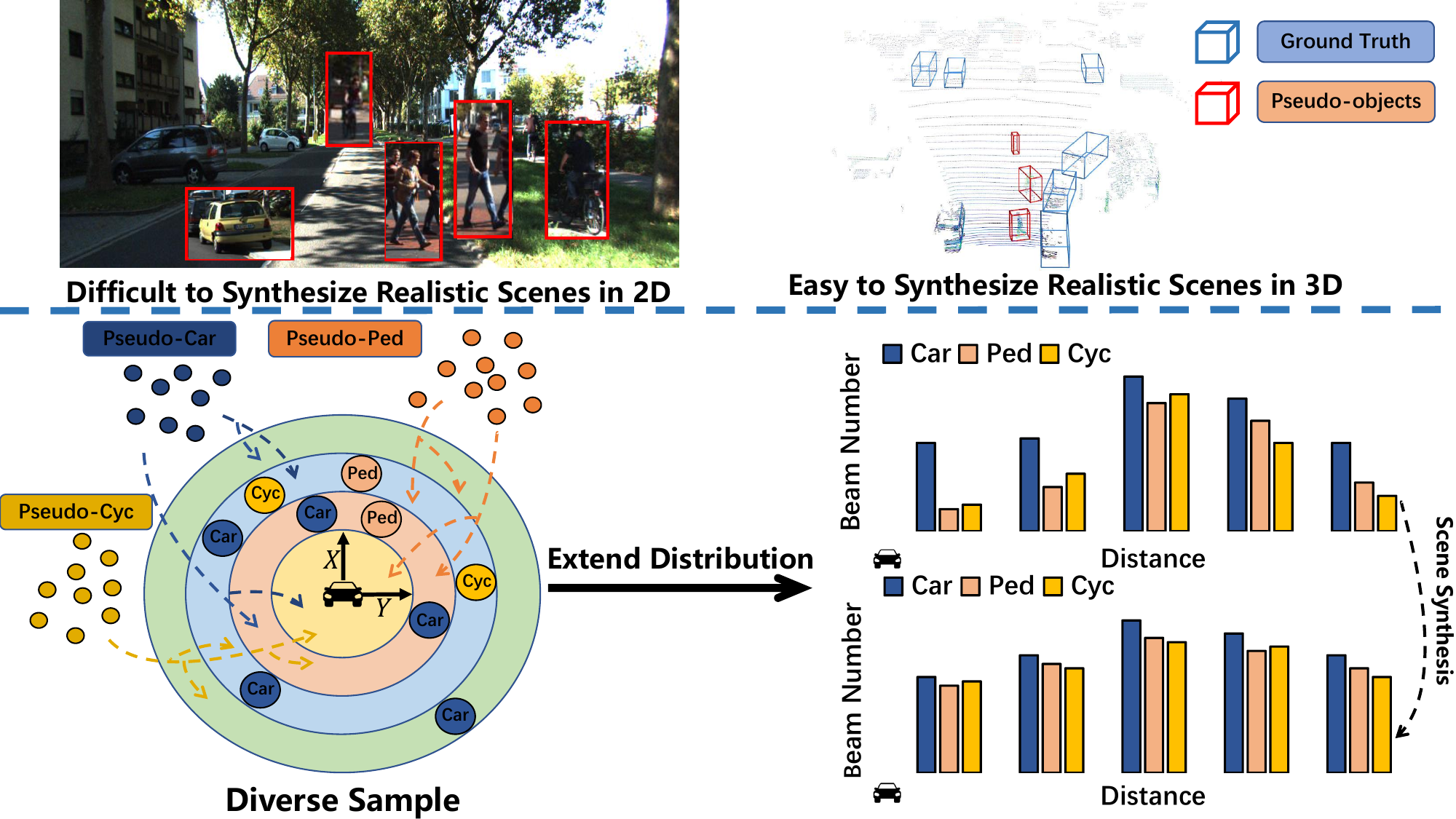}
\vspace{-7mm}
\caption{An illustration of constructing a synthetic 2D image sample and a 3D LiDAR sample. 
We exemplify the difficulty of constructing a 2D image and a 3D scene with objects from unlabeled data. 
It is difficult to synthesize a realistic 2D image yet easy to synthesize unlabeled objects at different positions to generate a diverse 3D sample to extend the laser beam distribution. }
\figvspace
\label{fig: intro}
\end{figure}

Semi-supervised learning methods usually generate pseudo-labels \cite{berthelot2019mixmatch,sohn2020fixmatch,lee2013pseudo} to propagate information from the labeled to the unlabeled data, rendering filtering noisy pseudo-labels a critical procedure. 
This motivates many semi-supervised object detection methods to develop more effective filtering strategies to obtain high-quality pseudo-labels with supervisory information. 
These methods typically design evaluation metrics \cite{liu2021unbiased,sohn2020fixmatch,wang20213dioumatch}, such as uncertainty \cite{liu2021unbiased}, confidence \cite{sohn2020fixmatch} or estimated IoU \cite{wang20213dioumatch}, and then filter the predictions according to the metric scores. 
However, it is difficult for manually designed evaluation metrics to accurately measure the quality of pseudo-labels, resulting in many false pseudo-labels to supervise the model and many accurate pseudo-labels discarded. 
Some existing semi-supervised 3D object detection methods \cite{wang20213dioumatch,yin2022semi,zhao2020sess} follow the pseudo-label-based 2D methods \cite{yang2022mix,liu2021unbiased,xu2021end,sohn2020simple,xu2023efficient} and achieve good results.
Still, most of them ignore the spatial characteristics of point clouds and cannot modify the distribution of data features, limiting the generalization ability of the trained model. 
Compared with 2D images, the point cloud scenes are not grid-structured and can be more easily synthesized by mixing two samples.
As shown in Figure \ref{fig: intro}, it is more difficult to construct realistic scenes with 2D images yet much easier with point clouds.

Motivated by this, we propose a hardness-aware scene synthesis (HASS) method to generate more diverse samples that can improve model generalization through scene synthesis.
We maintain an online pseudo-database containing both ground truth and pseudo-labels and progressively add the pseudo-labels that satisfy the threshold conditions generated in training.
We then randomly sample the foreground point clouds from the pseudo-database and concatenate them with the labeled point clouds to synthesize diverse samples.
To alleviate the influence of low-quality pseudo-labels, we employ harder (i.e., more difficult) synthetic samples to train the model when it is better learned~\cite{hdml} and maintain the pseudo-database progressively and adaptively.
We use a dynamic pseudo-database to progressively challenge the model by adding the pseudo-objects to the pseudo-database.
Extensive experiments on the widely used KITTI \cite{geiger2013vision} and Waymo \cite{sun2020scalability} datasets demonstrate that our HASS improves the performance of existing methods by a large margin.
We also provide an in-depth analysis of the performances of HASS under different settings to demonstrate the effectiveness of each module.

\section{Related work}

\textbf{Semi-Supervised 3D Object Detection.}
While conventional methods focused on semi-supervised 2D object detection from images~\cite{yang2022mix,liu2021unbiased,xu2021end,sohn2020simple,xu2023efficient,chen2021temporal,lin2021unreliable, chen2023mixed}, recent methods exploited the characteristics of point clouds and developed various techniques for 3D object detection to utilized additional unlabeled data~\cite{wang20213dioumatch,yin2022semi,zhao2020sess, yuan2024ad, wang2023not,park2022detmatch}.
SESS\cite{zhao2020sess} introduced a comprehensive perturbation strategy and used asymmetric data augmentation to improve model generalization.
3DIoUMatch \cite{wang20213dioumatch} designed a network to estimate locations for pseudo-label localization.
Proficient Teachers \cite{yin2022semi} obtained bounding box features via a pre-trained RoI network \cite{deng2021voxel} and enhanced the recall of pseudo-labels by manually devising Spatial-temporal Ensemble.

Recent work~\cite{leng2022pseudoaugment, liu2022ss3d, wu2022boosting} explored applying data augmentation to improve semi-supervised object detection.
For example, PseudoAugment \cite{leng2022pseudoaugment} devised data augmentation strategies based on pseudo-labels to fuse both labeled and pseudo-labeled data to mine unlabeled data.
SS3D \cite{liu2022ss3d} proposed a sparse-supervised method and designed an instance mining module with a filter to mine positive instances.
However, direct scene synthesis cannot achieve satisfactory results due to the low quality of pseudo-labels, which may corrupt the pseudo-database and mislead the model.
Considering this, we propose a HASS framework to progressively generate harder yet more accurate synthetic samples for training.
We further design a dynamic pseudo-database to improve the quality and diversity of scene synthesis.

\textbf{Semi-Supervised Learning.}
Semi-supervised learning (SSL) is designed to train models with few labeled data and abundant unlabeled data.
The mainstream SSL methods can be divided into two categories: consistency regularization \cite{sajjadi2016regularization,bachman2014learning,tarvainen2017mean,rasmus2015semi,zhang2021real} and pseudo-labeling \cite{berthelot2019mixmatch,sohn2020fixmatch,lee2013pseudo,wang2021semi}.
The consistency regularization methods applied different augmentations to the input samples and enforced the consistency of the model predictions between the augmented samples.
\cite{bachman2014learning} added perturbations to the model to enforce consistency.
Mean Teacher \cite{tarvainen2017mean} used an exponential moving average (EMA) teacher model with the same architecture as the student model to improve robustness.
Pseudo-labeling methods \cite{berthelot2019mixmatch,sohn2020fixmatch,lee2013pseudo,wang2021semi} explicitly generate pseudo-labels on unlabeled data for supervising, which can be regarded as a type of consistency regularization as it uses the predictions of one model to supervise the output of another model.
Differently, the proposed HASS method generates diverse synthetic data to improve the generalization.

\textbf{3D Object Detection.}
 Voxel-based representation~\cite{zhou2018voxelnet,lang2019pointpillars,yan2018second,ren2016three,yin2021center,chen2019fast,zhang2023simple} is a common point cloud processing method in deep learning and has been widely used in 3D object detection.
 VoxelNet \cite{zhou2018voxelnet} grouped point cloud data into voxels and used a voxel feature coding layer to obtain voxel features.
 PV-RCNN \cite{shi2020pv} integrated both 3D voxel Convolutional Neural Network (CNN) and PointNet-based \cite{qi2017pointnet} set abstraction to learn features.
 MV3D \cite{chen2017multi} is the first method to convert point cloud data into a BEV representation for 3D detection, which became popular among 3D object detection methods due to its high efficiency~\cite{chen2017multi,liang2019multi,simony2018complex,yang2018pixor}.
 However, 3D object detection is usually limited by the labeled training data which are laborious to annotate.
 Our method aims at effectively utilizing the unlabeled samples to improve the performance.

\section{Proposed Approach}

\begin{figure*}[tb]
\centering
\includegraphics[width=1\textwidth]{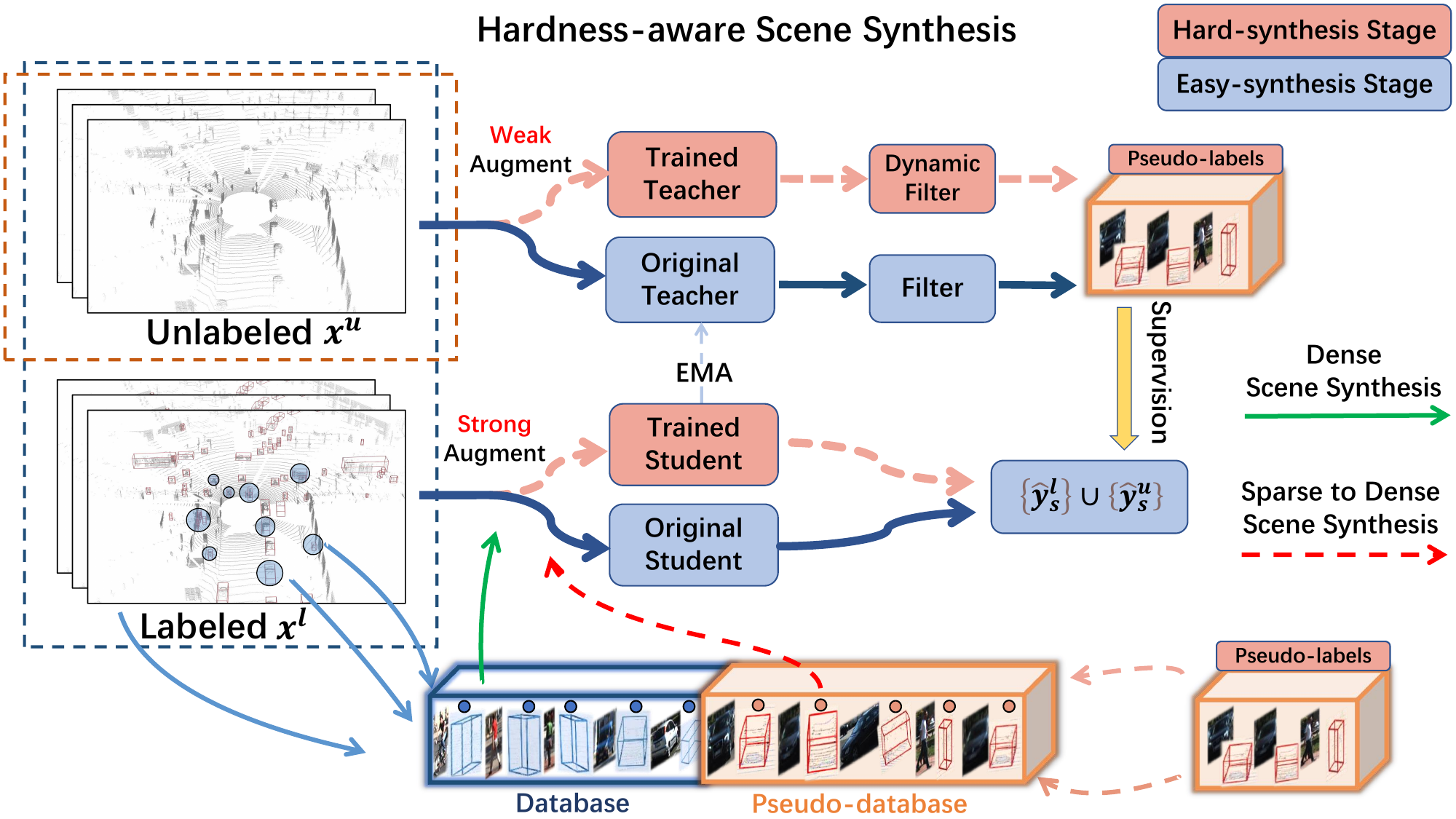}
\vspace{-7mm}
\caption{
Overview of the HASS framework.
The proposed architecture consists of two stages: (a) Easy Synthesis and (b) Hard Synthesis.
 We use blue arrows and light red dashed arrows to represent the easy-synthesis stage and hard-synthesis stage respectively.
 In the easy-synthesis stage, we only synthesize ground truth objects (light blue arrow), where the database is generated offline before training and contains only ground truth.
 As the training proceeds, the model is more tolerant of hard pseudo-labels.
 We maintain a pseudo-database and update the pseudo-database with the appropriate pseudo-labels in the hard-synthesis stage (light pink dashed arrow).
  The green arrow represents the dense synthesis strategy for the easy-synthesis stage, and the red dashed arrow represents the sparse to dense synthesis strategy for the hard-synthesis stage.
  }
\figvspace
\label{fig: Overview}
\end{figure*}

\subsection{Overview}\label{subsection:formulation}
Given a labeled point clouds sample $\boldsymbol{x}^l_i \in \mathbb{R}^{N\times4}$ containing a set of objects $y_i = \{y_{i1}, y_{i2},  \cdots , y_{ij}\}$ and each $y_{ij}$ containing a class code $\l_j$, bounding box parameters $b_j=\{c_j, d_j, h_j\}$. $c_j$, $d_j$, and $h_j$ represent the center coordinates, dimensions, and rotation angles of the bounding box $y_{ij}$, respectively. Furthermore, assume we have access to an unlabeled point clouds sample $\boldsymbol{x}^u_i \in \mathbb{R}^{N\times 4}$.

Fully-supervised learning only uses $\{\boldsymbol{x}^l_i\}_{i=1}^{N_l}$ as supervision, where $N_l$ is the number of labeled samples.
Semi-supervised learning aims to mine effective features from the unlabeled data $\{\boldsymbol{x}^u_i\}_{i=1}^{N_u}$ without reliable supervision, where $N_u$ is the number of unlabeled samples.
Pseudo-labeling-based semi-supervised detection predicts $\boldsymbol{\tilde{y}}_i$ from the $\boldsymbol{x}^u_i$ point clouds which contains a set of objects $\tilde y_i = \{\tilde y_{i1},\tilde y_{i2}, \cdots ,\tilde y_{ij}\}$.
Each $\tilde y_{ij}$ consists of class code $\tilde l_j$, $\tilde b_j=\{\tilde c_j, \tilde d_j, \tilde h_j\}$.
$\tilde y_i$ is not completely accurate but can reflect semantic features of $\boldsymbol{x}^u_i$ point clouds.
Pseudo-labeling-based semi-supervised detection methods usually leverage a teacher-student framework to propagate information in the form of pseudo-labels $\tilde y_i$.

The teacher model can aggregate information and produce a more accurate model than using the weights directly.
The predictions $\tilde y_i$ of the teacher model are more reliable.
We use the teacher model to predict $\{\boldsymbol{\tilde{y}}_i\}_{i=1}^{N_u}$:
\begin{equation}
    \{\boldsymbol{\tilde{y}}_i\} = f({x}^u_i, \theta_t), \forall i = 1, \cdots, N_u,
\end{equation}
where $\theta_t$ is the parameters of the teacher model at time t.
The final loss can be formulated as:
\begin{equation}
     L = L_{l}(\{\boldsymbol{x}^l_i\}_{i=1}^{N_l}, \{\boldsymbol{y}_i\}_{i=1}^{N_l}) + \lambda_{u}L_{u}(\{\boldsymbol{x}^u_i\}_{i=1}^{N_u},\{\boldsymbol{\tilde{y}}_i\}_{i=1}^{N_u}),
\end{equation}
where $\lambda_{u}$ is the unsupervised loss weight.

Most existing pseudo-labeling semi-supervised methods for 3D detection follow the semi-supervised 2D detection methods.
They usually define a pseudo-label quality evaluation metric, and then filter the pseudo-labels $\tilde y_i$ according to the metric.
Despite good performance, they ignore the unstructured nature of the point clouds, which are different from planar-like images that cannot reflect the actual geometry of objects.
The spatial structure of pixels is difficult to change, and point clouds faithfully reflect the distribution of the physical world.
Therefore, we can easily construct a point cloud scene with any foreground or background objects without rendering \cite{rao2021randomrooms}, facilitating efficient mining of unlabeled data $\{\boldsymbol{x}^u_i\}_{i=1}^{N_u}$.
In this paper, we propose a hardness-aware scene synthesis (HASS) framework to generate adaptive synthetic scenes to improve the generalization, as shown in Figure~\ref{fig: Overview}.

\begin{figure*}[tb]
\centering
\includegraphics[width=1\textwidth]{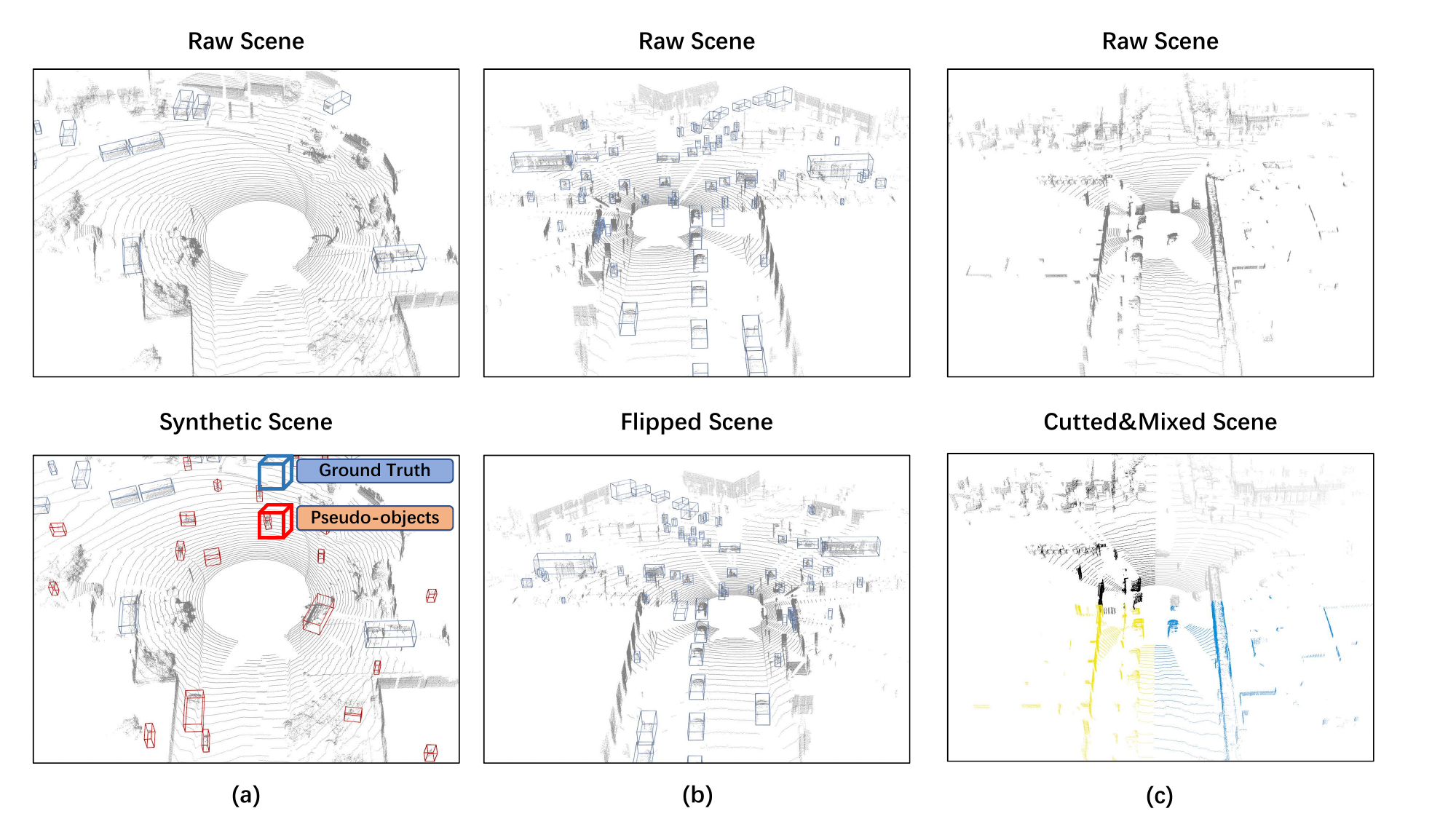}
\vspace{-7mm}
\caption{An illustration of the proposed (a) Scene Synthesis and (b) the Flip, (c) PointCutMix data augmentation methods.
(a): Blue bounding boxes represent existing ground truth boxes, and red bounding boxes represent synthetic objects.
Scene synthesis is to synthesize the foreground pseudo-labels from the pseudo-database on the labeled point clouds, where the synthetic scenes contain abundant unseen foreground information.
We avoid object collision to synthesize the correct scenes.
(b): Flip augmentation is the straightforward process of flipping the point cloud to obtain similar data.
(c): PointCutMix involves replacing parts of its own point cloud with point clouds from other scenes.
Point clouds of different colors represent data from different scenes.}
\figvspace
\label{fig:scene}
\end{figure*}

\subsection{Objects \& Background Composition}\label{subsection:Scene Synthesis}

We synthesize the labeled point clouds $\boldsymbol{x}^l_i$ and the foreground point clouds $\tilde p_j$ from the unlabeled samples $\{\boldsymbol{x}^u_i\}_{i=1}^{N_u}$ into a scene, as shown in Figure \ref{fig:scene}.
We then obtain a set of objects in $\boldsymbol{x}^l_i$:
\begin{equation}
    y_i = \{y_{i1}, y_{i2},  \cdots , y_{ij}, \tilde y_{i1},\tilde y_{i2}, \cdots ,\tilde y_{ik}\},
\end{equation}
where $y_i = \{y_{i1}, y_{i2},  \cdots , y_{ij}\}$ is manually labeled.
Given an unlabeled data set $\{\boldsymbol{x}^u_i\}_{i=1}^{N_u}$, we use the teacher model to predict each sample in the set to generate foreground pseudo-labels $\{\boldsymbol {\tilde y_i}\}_{i=1}^{N_u}$.
In addition, we maintain a pseudo-database $\tilde D$ of pseudo-labels $\tilde y_i$, and the qualified pseudo-labels will be added to the database.
We obtain a pseudo-database $\tilde D=\{\tilde y_{1},\tilde y_{2}, \cdots ,\tilde y_{n}\}$ that is continually updated and contains a lot of foreground information about unlabeled data $\{\boldsymbol{x}^u_i\}_{i=1}^{N_u}$.
Specifically, we concatenate foreground pseudo-labels $\tilde y_{n}$ from the pseudo-database $\tilde D$ with manually labeled objects $y_i$.
The synthesis of foreground objects $\tilde y_{n}$ on the labeled background $\boldsymbol{x}^l_i$ will not cause object collision, which ensures that the synthetic point clouds $\tilde{x}^l_i$ is the correct reflection of the real world as the annotation information specifies the location of all objects $y_i$ in the labeled scenes.
Before synthesizing a scene, we remove background point clouds $\boldsymbol{x}^l_i$ in the labeled scene to prevent collisions among the synthesized objects $\tilde y_{n}$.

While the existing methods~\cite{wang20213dioumatch} only uses ${x}^l_i$ with $y_i$, we synthesize powerful $\tilde{x}^l_i$ containing $\tilde y_i$ as input to the model.
The unsupervised loss is formulated as:
\begin{equation}
    L_{l}(\{\boldsymbol{\tilde x}^l_i\}_{i=1}^{N_l}, \{\boldsymbol{\tilde y}^l_i\}_{i=1}^{N_l}),
\end{equation}
where $\{\boldsymbol{\tilde y}^l_i\}_{i=1}^{N_l}$ are manual labels and the predictions of other temporal teacher models.
Compared to the other data synthetic methods \cite{chen2021shape, liu2019generative}, our proposed HASS has different synthesized content and is simpler to generalize.

\subsection{Hardness-Aware Scene Synthesis}\label{subsection:HASS}
\begin{figure}[tb]
\centering
\includegraphics[width=0.475\textwidth]{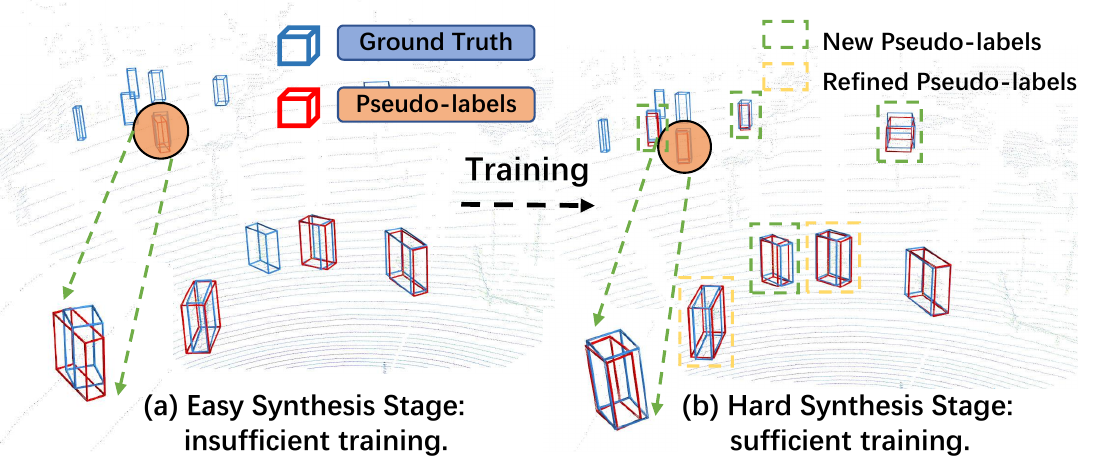}
\vspace{-7mm}
\caption{The visualization of detection with different models.
Blue boxes represent ground truth boxes and red boxes represent pseudo-labels.
We visualize the detection of teacher models generated in different epochs on the same sample.
(a) shows the low recall rate of the original teacher model at the beginning of training. The orange circle points out a false pseudo-label: a pedestrian identified as a cyclist.
(b) shows that the recall rate of the trained teacher model is higher than the original model.
The trained model predicts hard samples better than the original model.
}
\figvspace
\label{fig:det_cmp}
\end{figure}

Not all pseudo-labels $\tilde y_i$ can be used for effective scene synthesis.
Therefore, the proposed method has strict requirements on the quality of pseudo-labels $\tilde y_i$, and pseudo-labels $\tilde y_i$ produced by less trained teacher models are mostly of lower quality.
As shown in Figure \ref{fig:det_cmp}, low-quality pseudo-labels generated by the original model $\tilde y_i$ commonly contain only a fraction of the ground truth, which cannot sincerely reflect the geometry of ground truth, so the use of low-quality pseudo-labels $\tilde y_i$ can easily damage the precision of training data.
These low-quality pseudo-labels are hard to use as input synthetic samples for training.
This may cause the model to be trained in the wrong direction from the beginning, resulting in lower quality pseudo-labels $\tilde y_i$.

As training progresses, the model becomes better at handling hard pseudo-labels, and the trained teacher model generates higher-quality pseudo-labels $\tilde y_i$.
Therefore, the hard pseudo-labels can be synthesized in the input samples, and pseudo-labels $\tilde y_i$ generated by the trained teacher model are more suitable for scene synthesis.
We propose a hardness-aware scene synthesis to handle low-quality pseudo-labels $\tilde y_i$ from the trained teacher model, making them usable for scene synthesis.
The difference between the ``easy-synthesis" and ``hard-synthesis" stages lies in whether pseudo-labels are used for synthesis, as learning from pseudo-labels presents challenges for the model.
When to enter the ``hard-synthesis" stage depends on the level of training hardness for the model, and we proceed to the ``hard-synthesis" stage when the hardness of training samples sufficiently decreases.

Figure \ref{fig:det_cmp} shows the detection result of the same sample by different teacher models.
The recognition accuracy of the teacher model increases using more training samples.
We only add pseudo-labels $\tilde y_i$ generated by the trained teacher model to the pseudo-database $\tilde D$.
For the under-trained model, the difficult samples are more likely to produce low-quality pseudo-labels $\tilde y_i$.
For the better-trained model, the quality of pseudo-labels $\tilde y_i$ generated by predicting difficult samples increases and can be enough for scene synthesis.
Although there will inevitably be low-quality pseudo-labels, the trained model tolerates these hard low-quality pseudo-labels more than the original model.
In the easy-synthesis stage, we only use the manual labels in the ground database $ D= \{y_{i1}, y_{i2},  \cdots , y_{ij}\}$ for scene synthesis without adding the pseudo-labels $\tilde y_i$ generated by the original model to the pseudo-database $\tilde D$.
Only using these ground-truth labels $y_i$ for scene synthesis also accelerates the convergence of the model.
In the hard-synthesis stage, the model learns a large amount of data and can produce more accurate pseudo-labels $\tilde y_i$ from difficult samples.
The pseudo-labels $\tilde y_i$ generated by the trained model are then filtered according to the threshold and added to the pseudo-database $\tilde D$ to increase the diversity.

\subsection{Dynamic Pseudo-database}\label{subsection:Dynamic Pseudo-Database}
The estimated IoU \cite{wang20213dioumatch} and the confidence can accurately reflect the quality of the pseudo-labels $\tilde y_i$.
The relationship between pseudo-labels IoU and filtering threshold is given in Section \ref{exp:quality}.
If the filtering threshold is too high, a large amount of foreground information of unlabeled samples will be lost.
If the threshold for filtering is set too low, a low-quality pseudo-database will be obtained and the model will be trained on the wrong training data.
The density of scene synthesis is also crucial in different training stages.
 Dense scene synthesis is not meaningful for the model when the quality of the pseudo-database $\tilde D$ is not high.
 However, with a high-quality pseudo-database, sparse scene synthesis cannot achieve efficient mining of unlabeled data.

To solve these issues, we introduce a dynamic pseudo-database $\tilde D$ method. The threshold for filtering objects into the pseudo-database varies from high to low, while the scene synthesis density ranges from sparse to dense.
Moderate threshold filtering with either estimated IoU or confidence inevitably leads to low-quality pseudo-labels $\tilde y_i$ in the pseudo-database $\tilde D$.
At the beginning of adding pseudo-labels $\tilde y_i$ to pseudo-database $\tilde D$, the model recognition accuracy is still relatively low.
We then avoid misleading models with too many low-quality pseudo-labels by improving the entry threshold.
As the model becomes more tolerant of difficult samples, the quality of pseudo-labels $\tilde y_i$ with the same score increases.
To maintain a high learning efficiency, we lower the entry threshold to improve the diversity of pseudo-database $\tilde D$ information.

With the improvement of pseudo-database $\tilde D$ quality, it is necessary to adopt a flexible synthesis strategy.
We then adopt a sparse synthesis strategy at the beginning of updating the pseudo-database $\tilde D$.
We do not add too many foreground objects from the pseudo-database $\tilde D$ in each labeled point cloud to avoid more low-quality pseudo-labels $\tilde y_i$ and mitigate the influence of low-quality pseudo-labels $\tilde y_i$ on model training.
However, the quality of the pseudo-database $\tilde D$ will be higher when the model has learned a large number of hard samples at later stages of training.
More foreground objects from the pseudo-database $\tilde D$ can be synthesized in the annotated background point clouds, and we adopt a more dense synthesis strategy to sufficiently mine the features of unlabeled data.

Our method can be seen as a training strategy which leverages additional unlabelled data to improve the performance of 3D object detection in autonomous driving. 
The proposed framework can be applied to existing detectors during training to incorporate unlabelled data and does not introduce additional computation during inference.

\section{Experiments}

\begin{table*}[t] \small
\setlength{\tabcolsep}{4pt}
\renewcommand{\arraystretch}{1.05} 
    \begin{center}
    \caption{Comparisons with the 3DIoUMatch \cite{wang20213dioumatch} and DDS3D \cite{li2023dds3d} on the KITTI val set with different labeled ratios.
    The results are for moderate difficulty level evaluated by the mAP with 40 recall positions with a rotated IoU threshold of 0.7, 0.5, and 0.5 for the three classes, respectively. * denotes reproduced results.}
    \vspace{-3mm}
    \scalebox{1}[1]{
        \begin{tabular}{c|cccc|cccc|cccc}
        \hline
        & \multicolumn{4}{c|}{1\%} & \multicolumn{4}{c|}{2\%} & \multicolumn{4}{c}{10\%}\\
          \cline{2-13} \multirow{-2}{*}{Method} & Overall & Car & Pedestrian & Cyclist & Overall & Car & Pedestrian & Cyclist & Overall & Car &Pedestrian & Cyclist  \\
        \hline
        PV-RCNN~\cite{bachman2014learning} \cite{shi2020pv} & 44.3 & 73.8 & 28.2 & 30.8 & 53.3 & 76.4 & 39.6 & 43.9 & 62.7 & 79.7 & 49.1 & 59.2 \\ 
        3DIoUMatch~\cite{wang20213dioumatch} & 45.6 & 75.2 & 28.8 & 32.9 & 54.9 & 77.2 & 37.8 & 49.7 & 63.1 & 79.4 & 50.7 & 59.2 \\ 
        3DIoUMatch* & 47.8 & 76.0 & 30.7 & 36.8 & 59.0 & 78.6 & 45.3 & 53.0 & -- & -- & -- & -- \\ 
        DDS3D~\cite{li2023dds3d} & 49.8 & 76.0 & 34.8 & 38.5 & 60.7 & 78.9 & 49.4 & 53.9 & -- & -- & -- & -- \\ \cline{1-13}
        HASS & \textbf{51.4} & 75.9 & 31.0 & 47.2 & \textbf{61.7} & 79.0 & 44.5 & 61.7 & \textbf{65.9} & 79.8 & 52.8 & 65.1 \\ 
        \textit{Improvements} & \textit{+5.8} & \textit{+0.7} & \textit{+2.2} & \textit{+14.3} & \textit{+6.8} & \textit{+1.8} & \textit{+6.7} & \textit{+12.0} & \textit{+2.8} & \textit{+0.4} & \textit{+2.1} & \textit{+5.9} \\ \cline{1-13}
        \end{tabular}
        \label{kitti_main}
    }
    \end{center}
\tablevspace
\end{table*}

\begin{table*}[t] \small
\setlength{\tabcolsep}{3.5pt}
\renewcommand{\arraystretch}{1} 
    \begin{center}
    \caption{Comparisons with the 3DIoUMatch \cite{wang20213dioumatch} and Proficient Teachers \cite{yin2022semi} on the Waymo val set with different labeled ratios.
    mAP and mAPH under LEVEL 2 metric are used to evaluate the 3D object detection performance on the full validation set with 3D IoU thresholds of 0.7, 0.5, and 0.5 for Vehicle, Pedestrian, and Cyclist, respectively.}
    \vspace{-3mm}
    \scalebox{1.1}[1.1]{
        \begin{tabular}{c|c|c|c c c c}
        \hline
        & &  & \multicolumn{4}{c}{3D AP / APH @0.7 (LEVEL 2)}\\
          \cline{4-7} \multirow{-2}{*}{Label Amounts} & \multirow{-2}{*}{Model} & \multirow{-2}{*}{\textit{Improvements}} & Overall & Vehicle & Pedestrian & Cyclist    \\
        \hline
        \multirow{3}{*}{5\% ($\sim$ 4k Labels)} & 3DIoUMatch~\cite{wang20213dioumatch} & - / - & 48.85 / 43.91 & 52.41 / 51.26 & 46.05 / 35.85 & 48.10 / 44.63  \\ 
        & Proficient Teachers~\cite{yin2022semi} & +2.25 / +1.84 & 51.10 / 45.75 & 53.04 / 52.54 & 50.33 / 38.67 & 49.92 / 46.03   \\ 
        & HASS & +3.79 / +4.36 & 52.64 / 48.27 & 53.36 / 52.73 & 50.15 / 39.63 & 54.42 / 52.46  \\ \cline{1-7}
        \multirow{3}{*}{10\% ($\sim$ 8k Labels)} & 3DIoUMatch~\cite{wang20213dioumatch} & - / - & 53.39 / 48.25 & 56.43 / 55.65 & 51.11 / 40.10 & 52.63 / 49.01  \\ 
        & Proficient Teachers~\cite{yin2022semi} & +1.62 / +2.18 & 55.01 / 50.43 & 57.59 / 56.92 & 54.28 / 43.19 & 53.15 / 51.18   \\ 
        & HASS & +2.15 / +2.83 & 55.54 / 51.08 & 57.68 / 56.12 & 53.83 / 42.75 & 55.11 / 54.36  \\ \cline{1-7}
        \multirow{3}{*}{20\% ($\sim$ 16k Labels)} & 3DIoUMatch~\cite{wang20213dioumatch} & - / - & 57.2 / 52.35 & 59.37 / 58.22 & 55.59 / 44.38 & 56.63 / 54.46  \\ 
        & Proficient Teachers~\cite{yin2022semi} & +1.39 / +1.81 & 58.59 / 54.16 &  59.97 / 59.36 &  57.88 / 46.97 & 57.93 / 56.15   \\ 
        & HASS & +1.89 / +1.87 & 59.09 / 54.22 & 59.19 / 58.38 & 57.75 / 46.87 & 60.32 / 57.41  \\ \cline{1-7}

        \end{tabular}
        \label{waymo_main}
    }
    \end{center}
\tablevspace
\vspace{-3mm}
\end{table*}

\subsection{Experimental Settings}

In this section, we conducted various experiments on the widely used KITTI~\cite{geiger2013vision} and Waymo datasets~\cite{sun2020scalability} to verify the validity of our method.
We also provide in-depth experimental analysis and discussions about our framework.

\textbf{KITTI Dataset.} We experimented on 1\%, 2\%, and 10\% labeled data proportions.
In this experiment, we start updating the pseudo-database at 45th epoch and increasing the synthesis density increases from 5 to 15.

\textbf{Waymo Dataset.}
We experimented on labeled data proportions of 5\%, 10\%, and 20\%.
We proceeded to the hard-synthesis stage at the 15th epoch, where the density of synthesizing pseudo-objects increased from 10 to 30.

We provide more details about datasets, implementation, and evaluation metrics in Sections \ref{sec:dataset}, \ref{sec:imple}, and \ref{sec:metric}.

\subsection{Main Results}
We compared our HASS with DDS3D \cite{li2023dds3d} and 3DIoUMatch \cite{wang20213dioumatch} on the KITTI \cite{geiger2013vision} val set.
We also report comparisons between Proficient Teachers \cite{yin2022semi} and our proposed framework on the Waymo validation dataset.

\begin{table*} \small

\setlength{\tabcolsep}{12pt}
\renewcommand{\arraystretch}{1} 
{
    \caption{Ablation study on the effect of extending data distribution on 10\% labeled-only data.}
    \vspace{-6mm}
    \begin{center}

    \scalebox{1}[1]{
        \begin{tabular}{c|c|c|c|c|c|c|c|c|c}
        \hline
        & \multicolumn{3}{c|}{Car} & \multicolumn{3}{c|}{Pedestrian} & \multicolumn{3}{c}{Cyclist}\\
        \cline{2-10} \multirow{-2}{*}{Method} & Easy & Mod. & Hard & Easy & Mod. & Hard & Easy & Mod. & Hard  \\

        \hline

        PV-RCNN & 89.0 & 75.1 & 72.0 & 26.2 & 23.5 & 21.3 & 12.6 & 7.9 & 7.8 \\
        PV-RCNN-Extend & \textbf{91.2} & \textbf{79.4} & \textbf{76.6}  & \textbf{56.2} & \textbf{49.2} & \textbf{44.5}  & \textbf{77.1} & \textbf{55.7} & \textbf{52.3}\\
        \hline
        \textit{Improvements} & \textbf{+2.2} & \textbf{+4.3} & \textbf{+4.6}  & \textbf{+30} & \textbf{+25.7} & \textbf{+47.8}  & \textbf{+64.5} & \textbf{+47.8} & \textbf{+44.5}\\
        \hline

        \end{tabular}
    }
    \label{tabel:Extend}
    \tablevspace
    \vspace{-3mm}
\end{center}
}
\end{table*}

\begin{table}[t] \small
\setlength{\tabcolsep}{14pt}
\renewcommand{\arraystretch}{1} 
    \centering
    \caption{Comparisons of direct scene synthesis with different filtering thresholds on the KITTI val set.}
    \vspace{-3mm}
    \scalebox{1}[1]{
    \begin{tabular}{c|c|c|c}
        \hline
         {$\tau_{thresholds}$} &  Car & Pedestrian & Cyclist \\
        \hline
         Baseline & \textbf{79.4} & 49.2 & 55.2 \\
         $0.4_{Conf}$ & 77.3 & 48.6 & 65.5 \\
         $0.5_{Conf}$ & 77.7 & 50.6 & \textbf{62.1} \\
         $0.6_{Conf}$ & 78.0 & \textbf{51.4} & 61.8 \\
         \hline
    \end{tabular}
    }
    \label{Online}
    \tablevspace
    \vspace{3mm}
\end{table}

Table \ref{kitti_main} shows that the proposed method achieves competitive results on the KITTI dataset.
We see that our HASS achieves the best performance in all settings.
Specifically, in the 2\% 3DIoUMatch experiment reproduced by DDS3D\cite{li2023dds3d}, the Pedestrian detection performance differs from ours by 7.5.
Our overall performance in the 1\% proportion experiment is 1.6 higher than DDS3D~\cite{li2023dds3d}, and in the 2\% proportion experiment, it is 1.0 higher than theirs.
 Since the filter modules are difficult to filter the predictions of the original teacher model at the beginning, the predictions of the trained teacher model are relatively accurate, and the filter modules are able to filter out low-quality pseudo-labels.
  With the trained teachers, the filters are able to obtain high-quality pseudo-labels.
  The proposed method slightly improved performance for common samples such as Car and significantly improved performance for rare samples such as Pedestrian and Cyclist.
  The results verfies that hardness-aware scene synthesis produces data containing rare samples to extend data distribution and improve performance effectively.
 We also tested the inference speed of our model on a single NVIDIA GeForce RTX 3090 GPU and achieved an FPS of 5.9.
 Note that our method introduced no additional computations during inference.

The experimental results in Table \ref{waymo_main} indicate that our method also exhibits strong generalization on the large-scale Waymo dataset.
Our method shows significant performance improvement on a small proportion of labeled data.
In experiments with a 5\% proportion of labeled data, our method achieved an overall performance gain of 3.79/4.36 for AP/APH. In the 10\% and 20\% settings, the performance gains were 2.15/2.83 and 1.89/1.87, respectively.
We also maintain a leading position in overall performance gain Proficient Teachers \cite{yin2022semi}.

\subsection{Ablation and Analysis}

\textbf{\bf{Extending the Data Distribution}.}
We conducted experiments with fully-supervised PV-RCNN \cite{shi2020pv} on the KITTI dataset.
We followed gt-sampling \cite{yan2018second} to use a database composed entirely of labeled samples for scene synthesis.
Compared with the general method, scene synthesis achieves superior performance.
This demonstrates the effectiveness of using scene synthesis to extend the data distribution even without introducing unlabeled data.
Table \ref{tabel:Extend} shows that extending data distribution brings improvements of 4.3, 25.7, and 47.8 AP on moderate difficulty of Car, Pedestrian, and Cyclist, respectively.
We observe that the extended data distribution greatly improved the detection performance of rare samples such as Pedestrian and Cyclist.

 \begin{figure}[t]
\centering
\includegraphics[width=0.47\textwidth]{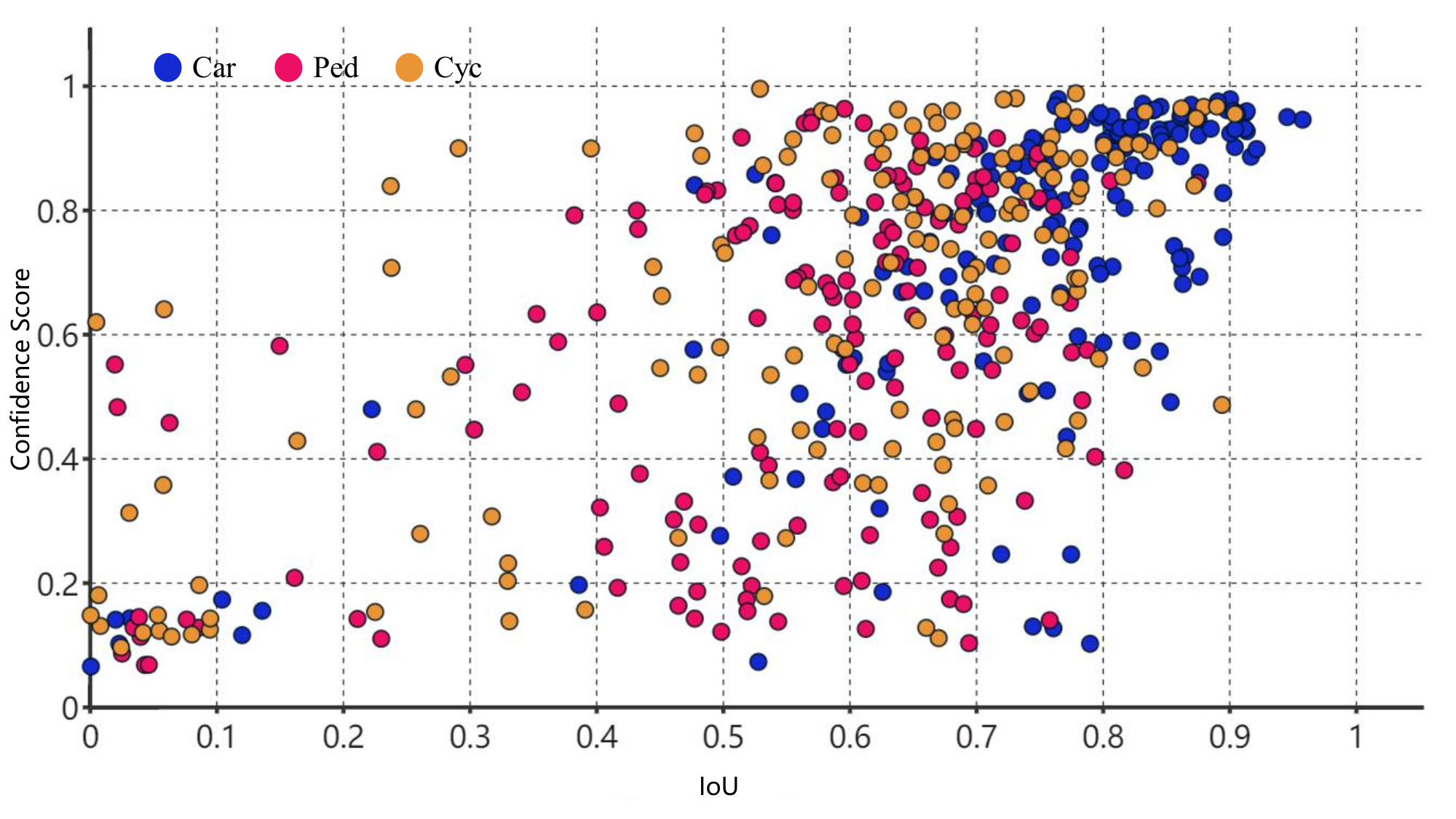}
\vspace{-3mm}
\caption{The visualization of the relationship between IoU and confidence of pseudo-labels.}
\label{fig: confi_iou}
\figvspace
\end{figure}

\textbf{\bf{Direct Scene Synthesis}.}
 We conducted direct scene synthesis experiments in different confidence filtering thresholds on the KITTI dataset.
 Based on 3DIoUMatch \cite{wang20213dioumatch}, we maintain an online pseudo-database.
 We experimented with thresholds of 0.4, 0.5, and 0.6, respectively.
 We observe an improvement in Pedestrian and Cyclist, but a decline in Car, as shown in Table \ref{Online}.
 We argue that the pseudo-labels produced by the original teacher filtering based on confidence threshold are not suitable for all types directly used for scene synthesis.

 In Figure \ref{fig: confi_iou}, we show the relationship between the IoU and the confidence of the pseudo-labels generated by the original teacher.
 Confidence cannot accurately reflect the IoU of pseudo-labels.
 We observe that many high-quality pseudo-labels have low confidence, so these pseudo-labels will be discarded in training, leading to a lack of unlabeled data information.
 However, there exist some low-quality pseudo-labels with high confidence.
 These low-quality pseudo-labels will be added to the pseudo-database during training and supervise the model.
 A high threshold (0.8) cannot completely filter out low-quality pseudo-labels (IoU \textless~0.6), and we cannot only rely on a high threshold to obtain high-quality pseudo-labels (IoU \textgreater~0.8).

 \begin{table}[t] \small
\setlength{\tabcolsep}{12pt}
    \centering
    \caption{Ablation study for exploring the effect of pseudo-labels produced by trained teacher model in offline scene synthesis.}
    \vspace{-3mm}
    \scalebox{1}[1]{
    \begin{tabular}{c|c|c|c}
         \hline
         Methods &  Car & Pedestrian & Cyclist\\
         \hline
         Baseline & 79.4 & 49.2 & 55.2 \\
         Offline & \textbf{79.9} & \textbf{52.5} & \textbf{66.3}  \\
         \textit{Improvements} & \textbf{+0.5} & \textbf{+3.3} & \textbf{+11.1}\\
         \hline
    \end{tabular}
    }
    \label{Offline}
    \tablevspace
    \vspace{1mm}
\end{table}

\begin{figure}[t]
\centering
\includegraphics[width=0.47\textwidth]{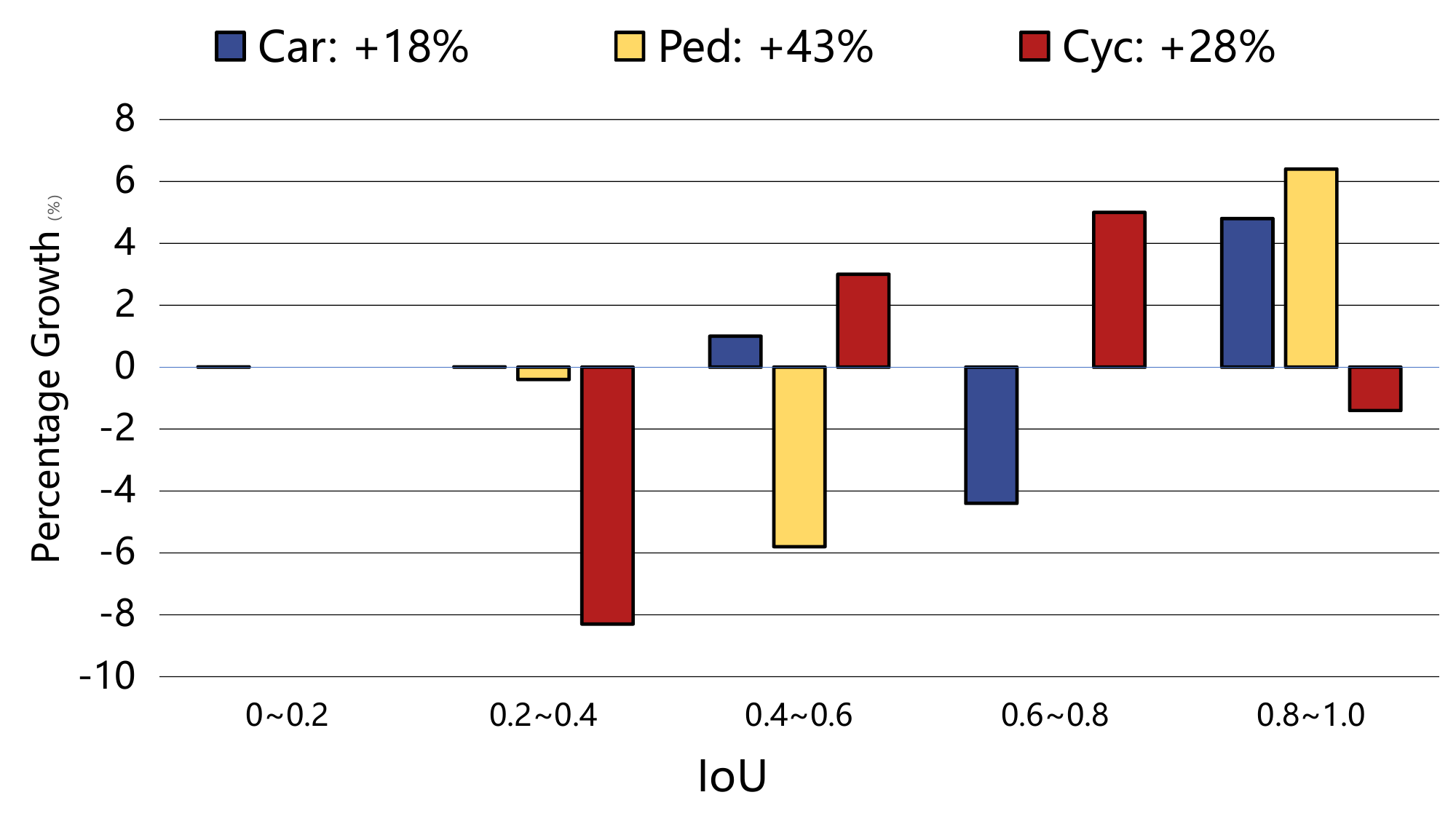}
\vspace{-4mm}
\caption{The quality improvement in the offline pseudo-database generated by a model trained for 60 epochs, compared to the directly generated pseudo-database filtered with a confidence threshold of 0.6.
The results did not count ground truth objects.}
\label{fig: offline_quality}
\figvspace
\end{figure}

\begin{figure*}[t]
    \centering
        \subfloat[Car\_Conf\label{fig:conf4}]{\includegraphics[width=0.33\textwidth]{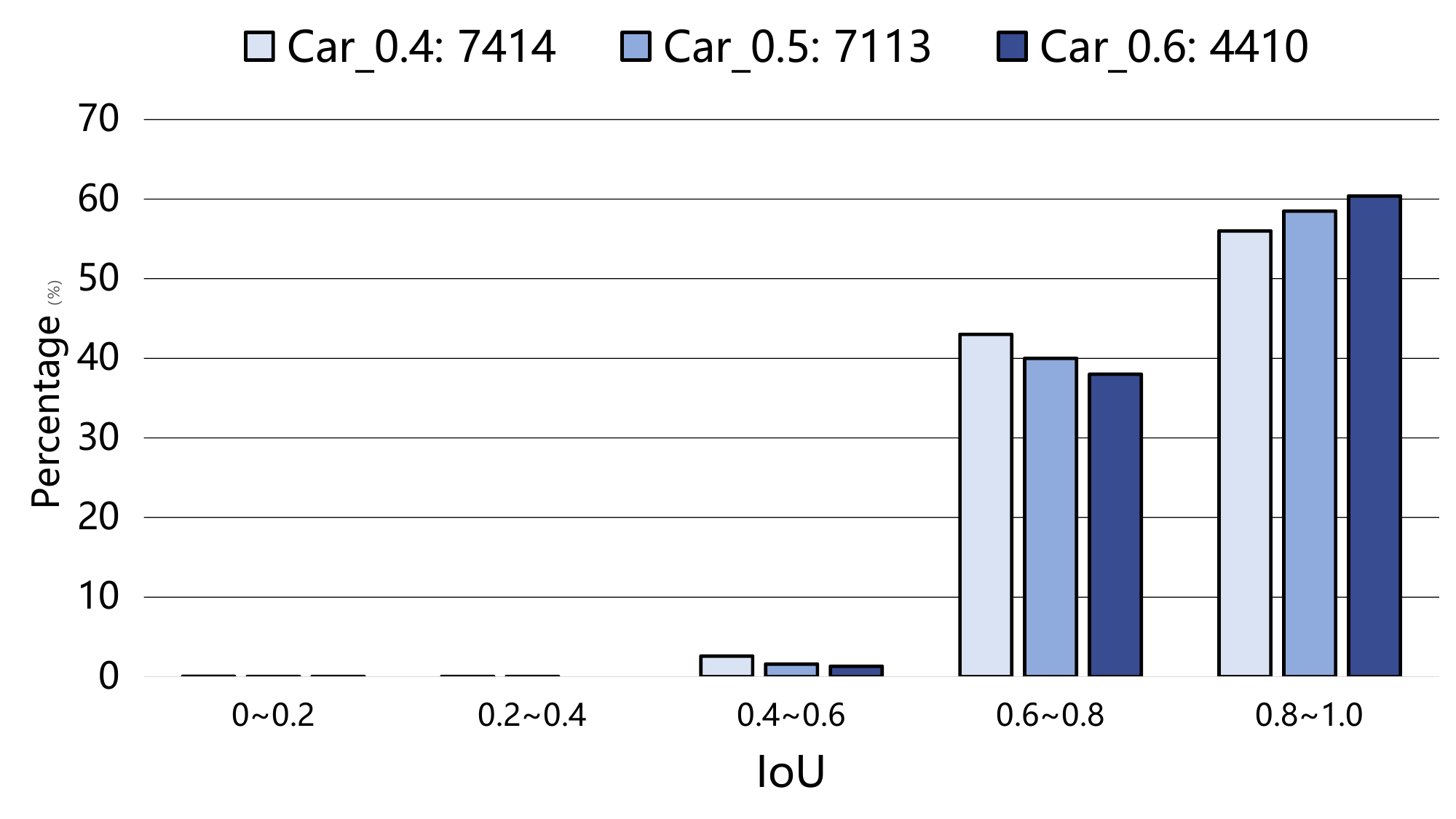}}
    \hfill 	
        \subfloat[Ped\_Conf\label{fig:conf5}]{\includegraphics[width=0.33\textwidth]{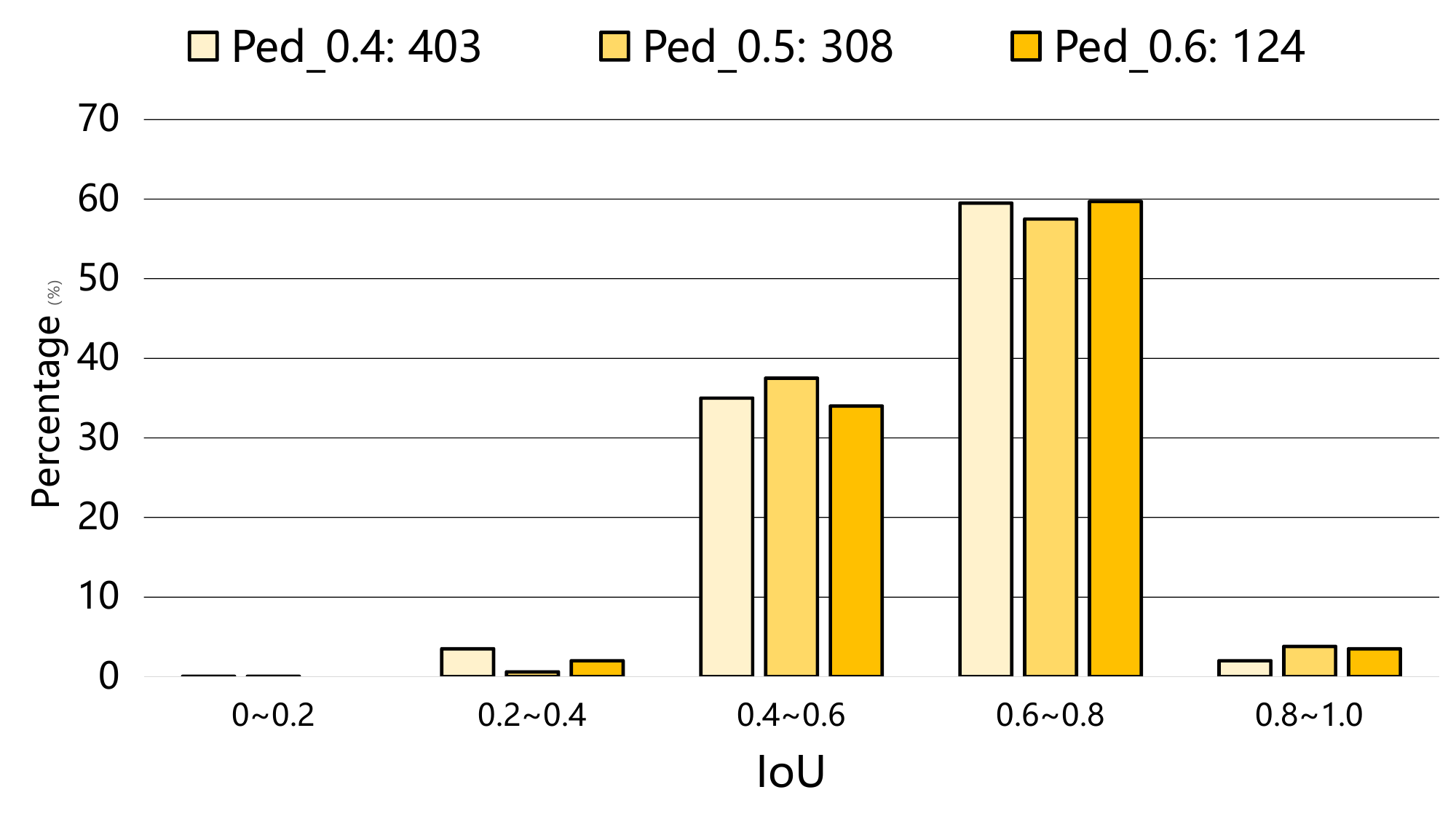}}
    \hfill 	
        \subfloat[Cyc\_Conf\label{fig:conf6}]{\includegraphics[width=0.33\textwidth]{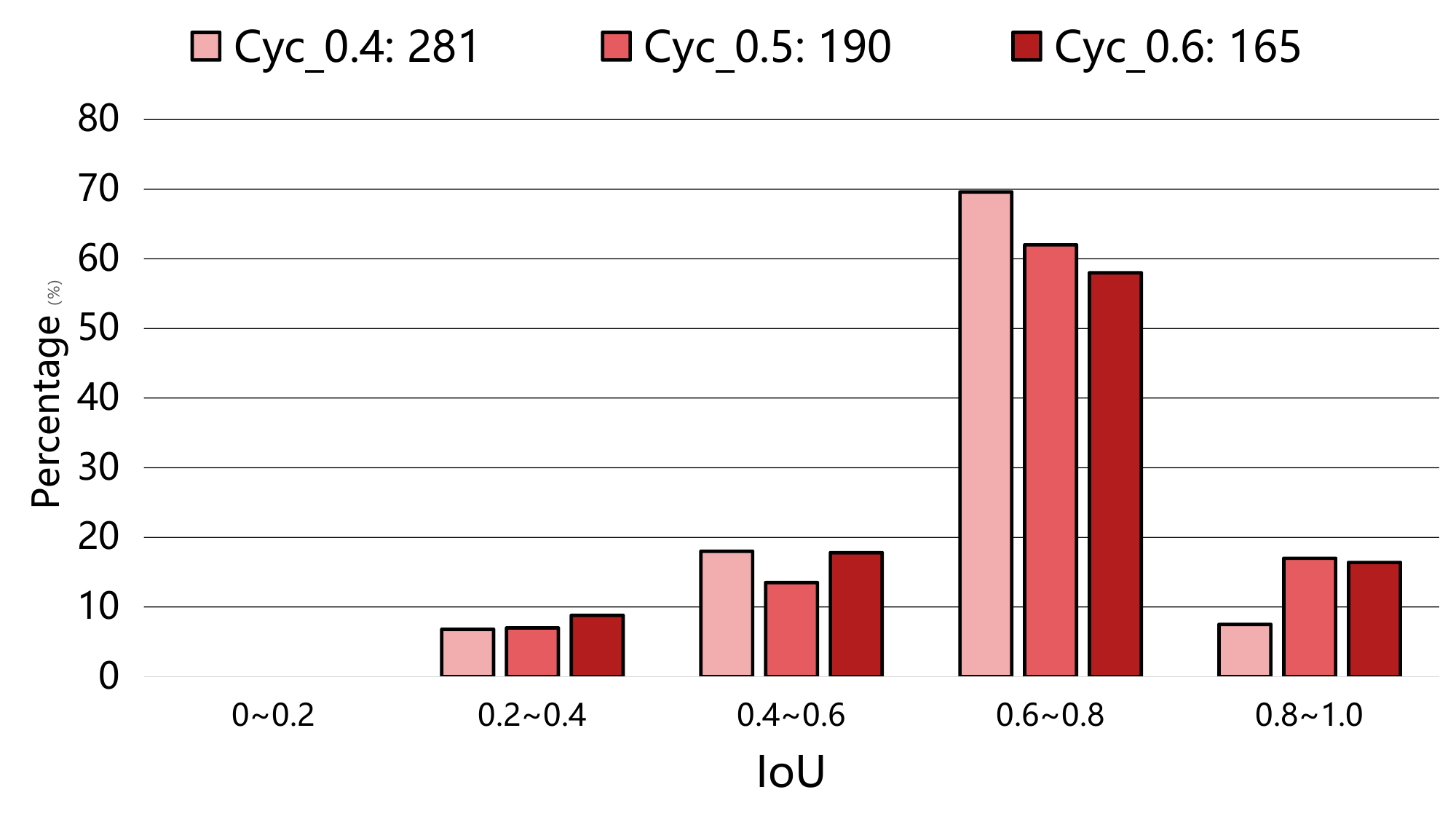}}
        \newline
        \subfloat[Car\_IoU\label{fig:IoU4}]{\includegraphics[width=0.33\textwidth]{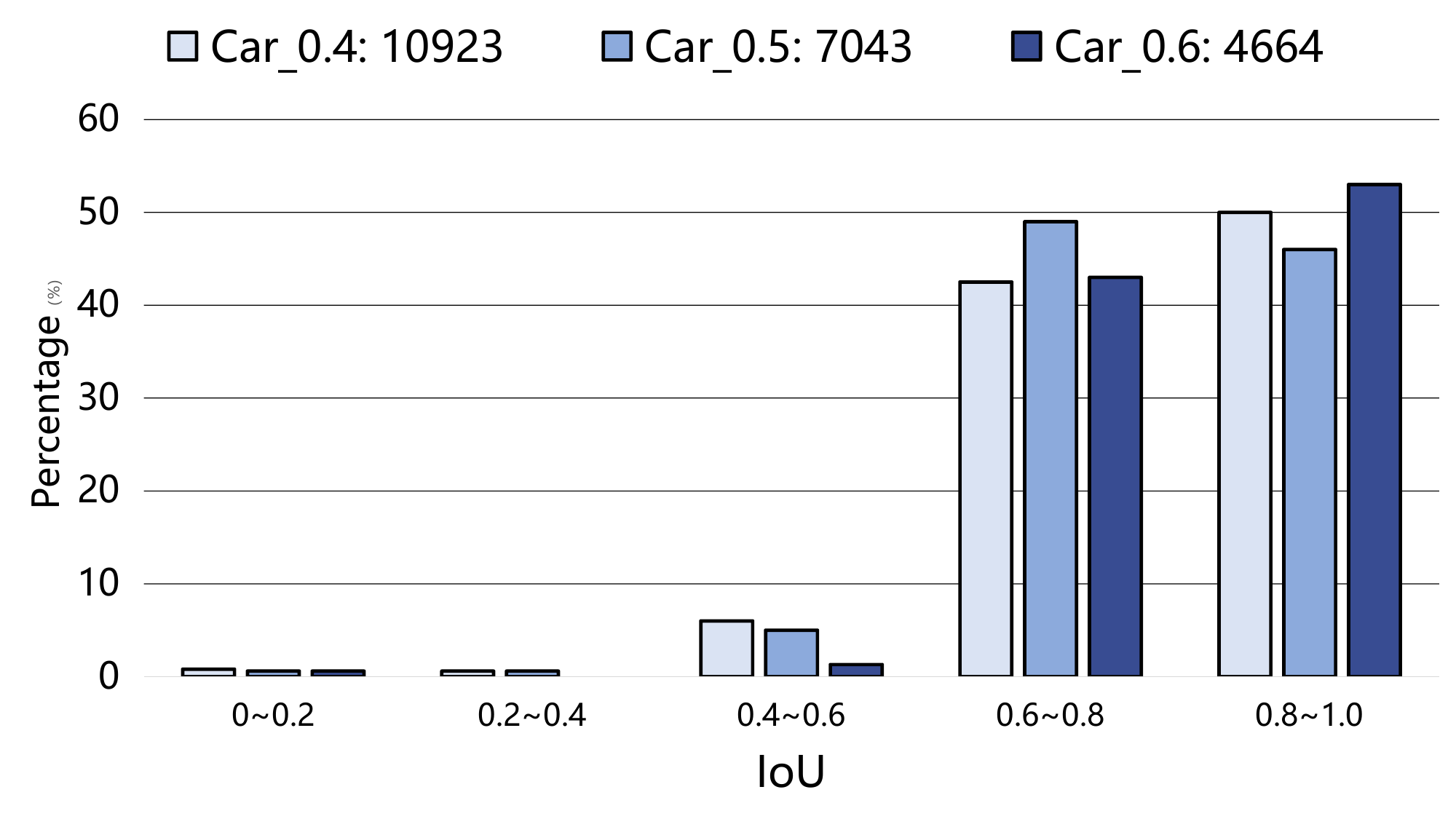}}
    \hfill
        \subfloat[Ped\_IoU\label{fig:IoU5}]{\includegraphics[width=0.33\textwidth]{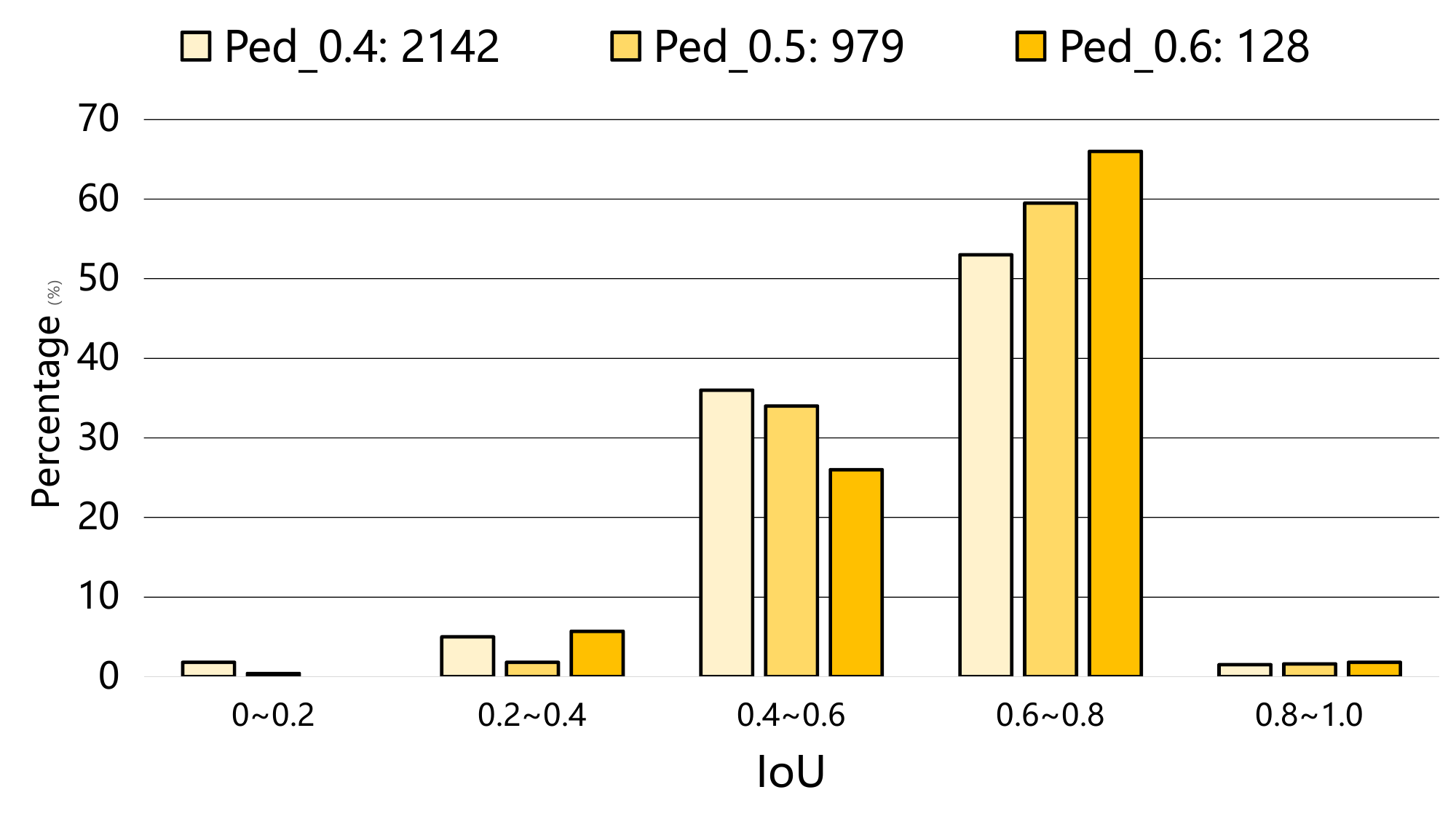}}
    \hfill
        \subfloat[Cyc\_IoU\label{fig:IoU6}]{\includegraphics[width=0.33\textwidth]{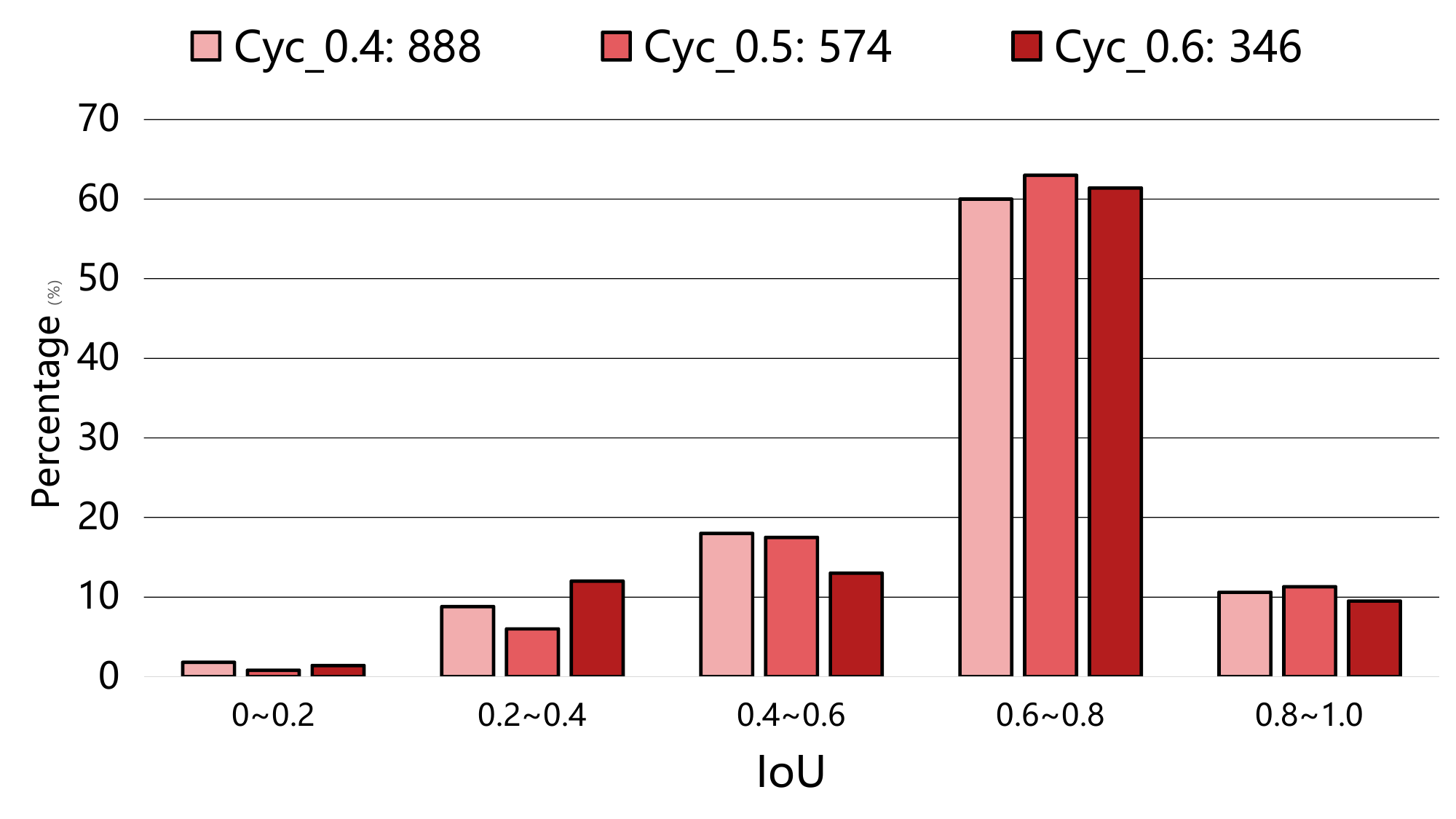}}
    \vspace{-3mm}
    \caption{Comparisons of pseudo-label quality with different filtering methods and thresholds.
    }
    \label{quality}
    \figvspace
    \vspace{2mm}
\end{figure*}

\begin{table*}[t] \small
\setlength{\tabcolsep}{4.3pt}
\renewcommand{\arraystretch}{1} 
    \begin{center}
    \caption{Ablation study of semi-supervised on 10\% labeled data and full unlabeled data of Waymo and KITTI datasets. T
    The KITTI results are for moderate difficulty level evaluated by the 3d mAP with 40 recall positions.
    We adopt the AP under LEVEL 2 metric to evaluate the 3D object detection performance on the Waymo dataset.
    }
    \vspace{-3mm}
    \scalebox{1.0}[1.0]{
        \begin{tabular}{c|ccc|ccc|c|ccc}
        \hline
         \multirow{2}{*}{Exp.} & \multicolumn{3}{c|}{Synthesized Density} & \multicolumn{3}{c|}{Hardness-aware level} & \multirow{2}{*}{Threshold} & \multicolumn{3}{c}{KITTI 10\% / Waymo 10\%}\\
         
         \cline{9-11}
        & sparse. & dense. & sparse to dense. & 25th/5th & 35th/10th & 45th/15th & & Car / Vehicle & Pedestrian & Cyclist \\
        \hline
        Baseline & & & & & & & & 79.4 / 56.43 &  49.2 / 51.11 &  55.2 / 52.63 \\ 
        (a) & \checkmark & & & & & \checkmark & & 79.5 / 56.85 & 52.9 / 53.36 &  61.1 / 54.51  \\ 
        (b) & & \checkmark & & & & \checkmark & & 79.4 / 56.79 & 49.8 / 52.82 & 57.2 / 54.92  \\ 
        \hline
        (d) & & & \checkmark & \checkmark & & & & 79.2 / 55.89 & 50.2 / 51.16 & 58.3 / 53.25 \\ 
        (e) & & & \checkmark & & \checkmark & & & 79.4 / 57.32 & 51.0 / 53.58 & 61.5 / 54.75  \\ 
        (f) & & & \checkmark & & & \checkmark & & 79.6 / 57.66 & 52.9 / 53.78 & 64.2 / 55.01  \\ 
        \hline
        (g) & & & \checkmark & & & \checkmark & \checkmark & \textbf{79.9 / 57.75} & \textbf{53.5 / 54.77} & \textbf{64.8 / 55.31} \\ \hline
        \end{tabular}
        \label{ablation}
    }
    \end{center}
\tablevspace
\vspace{-2mm}
\end{table*}

\begin{figure}[tb]
\centering
\includegraphics[width=0.47\textwidth]{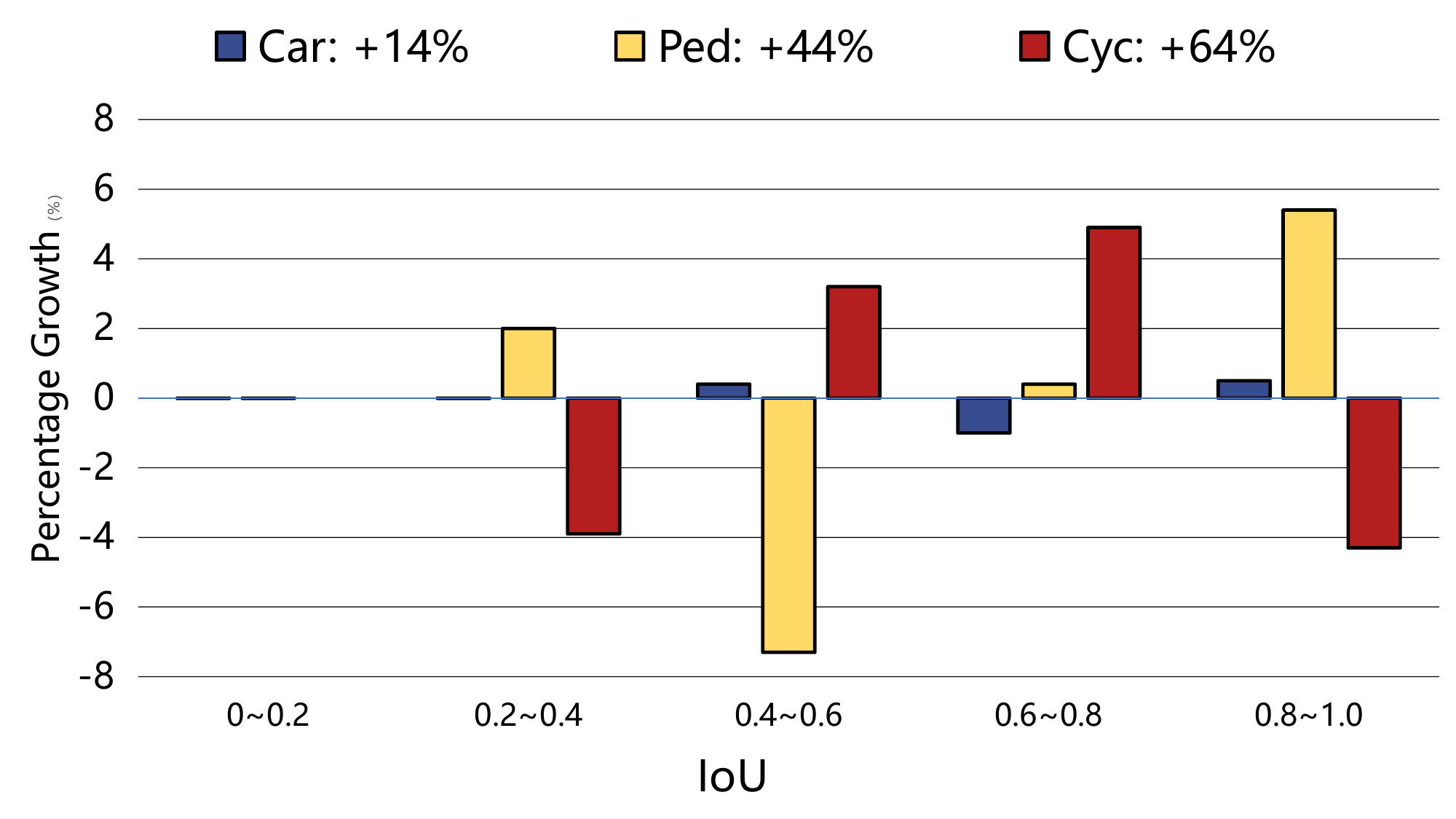}
\vspace{-4mm}
\caption{Ablation of the dynamic pseudo-database compared to the directly generated pseudo-database filtered with a confidence threshold of 0.6. The results did not include ground truth objects.
}
\figvspace
\label{fig: dynamic_quality}
\end{figure}

\textbf{Effect of the Dynamic Pseudo-database.}
We verify the effectiveness of the pseudo-labels produced by trained teacher filtering based on the confidence threshold for direct scene synthesis.
We maintain an offline pseudo-database generated by the fully trained 60 epochs teacher traverse unlabeled data based on a 0.6 confidence threshold.
Pseudo-labels in the offline database are not updated during training.
Table \ref{Offline} shows that offline scene synthesis outperforms 3DIoUMatch baseline by 0.5, 3.3, and 11.1 AP on Car, Pedestrian, and Cyclist, respectively.
This advocates the trained teacher model instead of the original teacher model to produce a pseudo-database.
Figure \ref{fig: offline_quality} shows the quality improvement of the offline pseudo-database.
We observe that the quality and diversity of the offline pseudo-database are better than the pseudo-database generated by the original model based on a 0.6 threshold.

 \textbf{\bf{Evaluation of Different Filtering Methods}.}\label{exp:quality}
 We compare the pseudo-label quality generated by different filtering methods and thresholds.
 We conducted pseudo-label filtering experiments on 10\% labeled data based on confidence \cite{shi2020pv} and estimated IoU \cite{wang20213dioumatch} on the KITTI dataset.
 As shown in Figure \ref{quality}, the number of pseudo-labels decreases with a higher threshold and a larger proportion of IoU.
IoU-based and confidence-based filtering has worked well for both car and cyclist pseudo-labels.
With similar quality, the IoU-based method gets more cyclists, while the quality of cars filtered by the confidence-based method is better than that of the IoU-based method.
The IoU-based method better filters Pedestrian pseudo-labels with low quality, but it does not achieve the expected results.
Pedestrian pseudo-labels with IoU of less than 0.6  still account for about 35\% after 0.5 IoU-based threshold filtering.
After filtering based on a 0.6 IoU-based threshold, the proportion of low-quality pedestrian pseudo-labels is reduced, but the number is also greatly reduced.
There are still many low-quality Pedestrian pseudo-labels (IoU \textless~0.6 account for more than 30\%) filtered by the two methods, which is related to the principle of estimating IoU.
The points of the pedestrian are like a cylinder, and the predicted bounding box is a cuboid.
The output estimated IoU of the module is likely to be similar for the bounding box with different rotation angles.
Both methods filter Car well, and high-quality pseudo-labels (IoU \textgreater~0.8) account for nearly 50\%.
However, the detection performance of the car declines in the experiment of direct scene synthesis.
We argue that the quality of pseudo-labels should be higher for such common objects as cars.

\begin{figure*}[t!]
\centering
\includegraphics[width=1\textwidth]{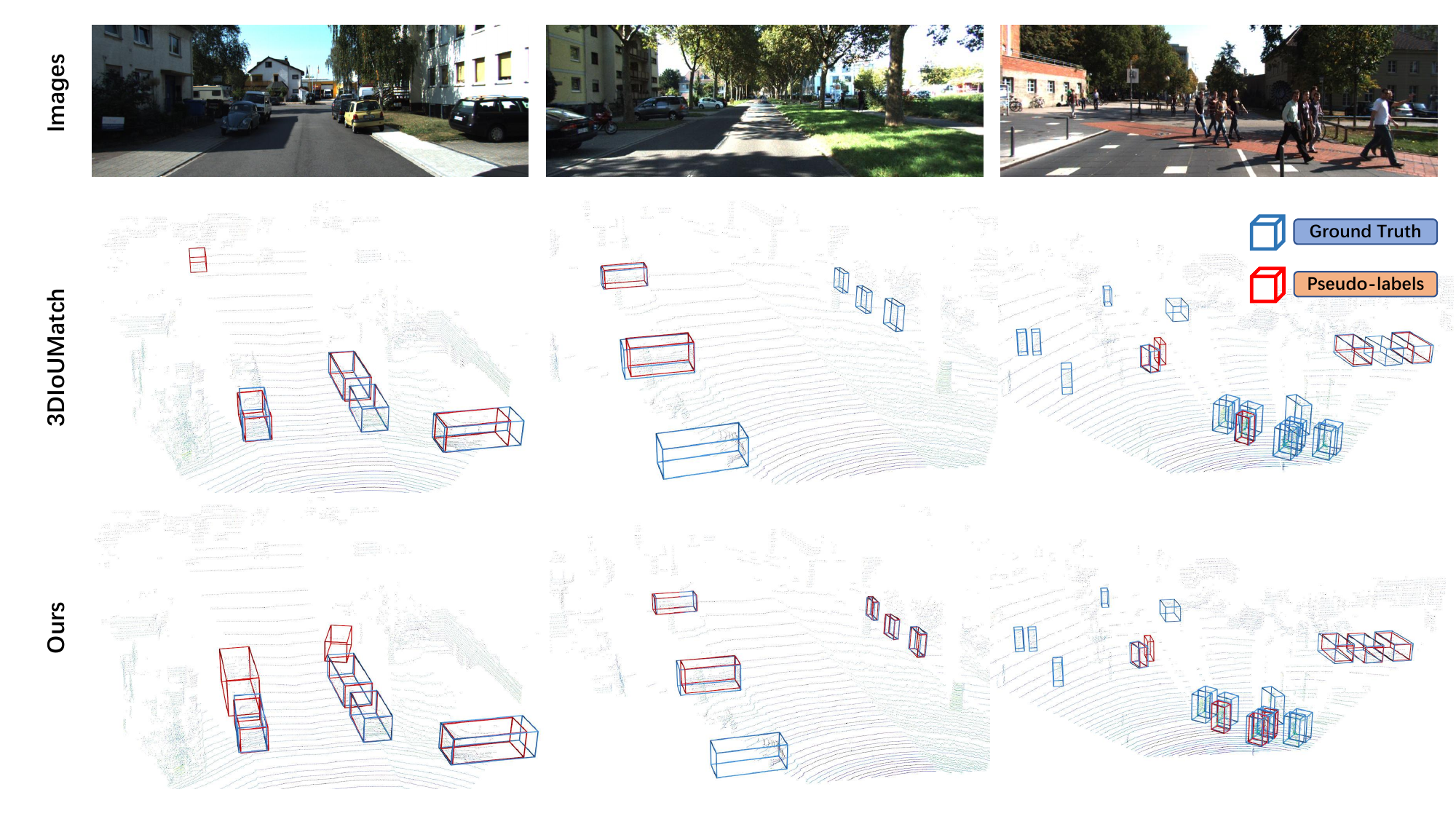}
    \vspace{-9mm}
\caption{Qualitative results of 3D object detection on KITTI with the 10\% labeled set.
Blue and red bounding boxes denote ground truths and predictions, respectively.}
\figvspace
\label{visualization}
\end{figure*}

 \textbf{\bf{Ablation Study of Dynamic Pseudo-database}.}
We compared hardness-aware scene synthesis using different synthesis strategies on the KITTI and Waymo datasets.
In the KITTI dataset experiments, the sparse synthesis strategy is to synthesize 5 objects per batch during the whole scene synthesis stage, while the dense strategy is to synthesize 15 objects.
The number of synthesized samples increases with the model learning process.
 Scene synthesis from sparse to dense is dynamically adjusted the synthesized density throughout the synthesized phase from 5 to 15.
 In the Waymo dataset, each sample contains more foreground objects compared to the KITTI dataset.
 Therefore, in the Waymo experiments, we doubled the number of synthesized pseudo-objects compared to that of the KITTI experiments.
 Table \ref{ablation} shows that the sparse-to-dense synthesis strategy achieves better results, which is active when the pseudo-database samples are more accurate and conservative when they are less accurate.
 The filtering threshold of experiments without dynamic threshold strategy is 0.5 and the dynamic filtering thresholds on the KITTI dataset and Waymo dataset were changed from 0.6 to 0.4 and from 0.8 to 0.4, respectively.
 Figure \ref{fig: dynamic_quality} shows that dynamic pseudo-database achieves higher quality and diversity in contrast to Figures \ref{fig:conf4}, \ref{fig:conf5}, and \ref{fig:conf6}.

\textbf{\bf{Ablation Study of Hardness-aware Level}.}
We tested the effect of hardness-aware on performance on the KITTI and Waymo datasets.
We compared the effect of scene synthesis based on the pseudo-database produced by different trained models.
The total training process for the KITTI and Waymo dataset experiments is terminated at the 60th epoch and the 30th epoch, respectively.
The three groups of experiments on the KITTI and Waymo datasets start updating the pseudo-database at the 25th and 5th epochs, respectively, with intervals of 10 epochs and 5 epochs.
Table \ref{ablation} shows that the hardness-aware scene synthesis method based on a pseudo-database produced by the 45th or 15th epoch-trained model achieves the best performance.
This is because the better-trained models can generate samples that are easier to learn with scene synthesis relying on the trained teacher.

\textbf{\bf{Visualizations}.}
We provided qualitative comparisons with 3DIoUMatch \cite{wang20213dioumatch} in Figure~\ref{visualization}.
We see that HASS produces pseudo-labels with better quality.
This is because we employ the hardness-aware scene synthesis method and the dynamic synthesis strategy, which gradually increases the density of scene synthesis to alleviate the impact of the initial pseudo-database on synthetic samples.

\section{Conclusion}
In this paper, we have presented a hardness-aware scene synthesis (HASS) framework for semi-supervised 3D detection. 
We propose to mine unlabeled data to generate diverse data to supplement training and extend the data distribution to improve the generalization ability. 
As scene synthesis requires high-quality pseudo-labels, we further use the pseudo-labels generated by the trained model instead of the original teacher for scene synthesis to reduce the number of low-quality pseudo-labels. 
We have adopted a sparse-to-dense scene synthesis strategy to produce more high-quality pseudo-labels while reducing the impact of low-quality pseudo-labels. 
We have conducted extensive experiments on the widely used KITTI and Waymo datasets to demonstrate the superiority of our framework and provided an in-depth analysis of the effectiveness of each module.

\appendix
\section{More Related Work}

\textbf{Semi-Supervised Object Detection.}
The ability to learn from additional unlabeled data is appealing for 2D object detection~\cite{yang2022mix,liu2021unbiased,xu2021end,sohn2020simple,xu2023efficient,chen2021temporal,lin2021unreliable, chen2023mixed}.
 STAC \cite{sohn2020simple} proposes an SSOL framework based on self-training and augmentation-driven consistency regularization.
MixPL\cite{chen2023mixed} addressed the issue of failing to detect small and tail objects for conventional image-based methods by using the mixup strategy.
Unbiased Teacher \cite{liu2021unbiased} followed the student-teacher mutual learning paradigm to address the pseudo-labeling bias issue.
To balance the effect of pseudo-labels, Soft Teacher \cite{xu2021end} proposed a soft teacher mechanism where the classification loss of each unlabeled bounding box is weighed by its classification score.

\textbf{Data Augmentation.}
Deep learning tasks often utilize small perturbations such as scaling, jittering, points dropout, flipping, and rotation to enhance the diversity of the training set~\cite{wang2021pointaugmenting,li2020pointaugment,zhang2022pointcutmix,yun2019cutmix}.
The benefits of these small perturbations diminish as datasets become larger, motivating the exploration of more effective data augmentation strategies.
CutMix \cite{yun2019cutmix} proposed to crop an image and blend patches from other images to create a new image sample for training.
PointCutMix \cite{zhang2022pointcutmix} extended the concept of CutMix by establishing correspondences between point clouds and then swapping them.
PointAugment \cite{li2020pointaugment} automatically selected data augmentation strategies (rotation, scaling, and translation) by iteratively narrowing down the search space in an end-to-end manner.
PointAugmenting \cite{wang2021pointaugmenting} further enhanced 3D object detection by decorating point clouds with point-wise CNN features extracted by pretrained 2D image detection models.

Data augmentation generally performs well in tasks with labeled data to prevent overfitting.
It involves adding noise to existing labeled samples, generating similar but not entirely identical samples.
Differently, our approach involves synthesizing a new sample using data from unseen unlabeled scenes, where the key challenge is to ensure data authenticity due to the quality of pseudo-labels.
We address this by using the hardness-aware adaptation to challenge the model with samples of adaptive difficulty.

\section{Datasets} \label{sec:dataset}
\textbf{KITTI Dataset.}
The KITTI dataset contains 7,481 scenes as training samples and 7,518 scenes as testing samples.
The 7,481 training samples were divided into a training set containing 3712 samples and a validation set containing 3769 samples.
KITTI contains point clouds of eight types of objects.

\textbf{Waymo Dataset.}
The Waymo dataset is composed of 798 sequences for training and 202 sequences for validation, and each sequence consists of approximately 198 frames of data.
Importantly, it provides abundant foreground information for nearly every scenario and offers annotations encompassing the entire scene rather than just specific viewpoints.

\section{Evaluation Metrics} \label{sec:metric}
According to the conventional KITTI experiments, we report the mAP for 3D boxes with 40 recall positions.
The performances on the Car, Pedestrian, and Cyclist classes are evaluated with IoU thresholds of 0.7, 0.5, and 0.5, respectively.
We use the 3DIoUMatch \cite{wang20213dioumatch} as our baseline.
We use mAP and mAPH under the LEVEL 2 metric to evaluate the 3D object detection performance in experiments on the Waymo dataset with 3D IoU thresholds of 0.7, 0.5, and 0.5 for "Vehicle," "Pedestrian," and "Cyclist," respectively.

\section{Implementation Details} \label{sec:imple}
\textbf{KITTI Dataset.}
Following existing methods \cite{wang20213dioumatch}, we adopted PV-RCNN \cite{shi2020pv} as our detector.
We conducted pre-training on 1\%, 2\% labeled-only data and increased the number of labeled data traversal of each epoch to 10 times that of PV-RCNN \cite{shi2020pv}.
Before pre-training, we initialized a database of labeled data points with ground truths \cite{yan2018second}.
During the easy-synthesis stage, the database remains unchanged, ensuring that all objects within it are entirely manually annotated.
In the hard-synthesis stage, at the end of each epoch, we added filtered pseudo-labels to the database.
For the pre-training stage, the detector was optimized by an ADAM optimizer with an initial learning rate of 0.01 for 80 epochs.
For the semi-supervised learning stage, 98 \% or 99 \% of training data are unlabeled data and the rest are labeled data.
We trained the detector for 60 epochs with the same training setting as pre-training.
We set the batch size as 2 (labeled: unlabeled, 1:1).
We adopted the same data pre-processing strategy as PV-RCNN.
Due to the randomness of 1\% and 2\% data proportions, we randomly selected three groups in proportion from the training set and presented the average results of the three groups.
Additionally, we expanded the evaluation of our method on the 10\% labeled data.
Unlike the 1\% and 2\% labeled data, we did not increase the number of traversal of labeled data.

\textbf{Waymo Dataset.}
Following existing works \cite{yin2022semi}, we evenly divided the 158K training samples into labeled and unlabeled sets.
We randomly sampled 5\%, 10\%, and 20\% sequences from the labeled set as labeled data.
The training data consists of sampled labeled data and the full unlabeled set.
The pre-training process involved traversing the labeled data only once per epoch.
For semi-supervised training, we set the batch size to 4 (labeled : unlabeled, 1:3), while keeping the other configurations the same as the KITTI.
 Both the pre-training and semi-supervised training phases consisted of 30 epochs.

{
    \small
    \bibliographystyle{ieeenat_fullname}
    \bibliography{main}
}

\end{document}